\documentclass{article}

\PassOptionsToPackage{numbers, compress}{natbib}

    \usepackage[preprint]{neurips_2020}

\usepackage[utf8]{inputenc} 
\usepackage[T1]{fontenc}    
\usepackage{hyperref}       
\usepackage{url}            
\usepackage{booktabs}       
\usepackage{amsfonts}       
\usepackage{nicefrac}       
\usepackage{microtype}      
\usepackage{amssymb}
\usepackage{amssymb}
\usepackage{xcolor}
\usepackage{amsfonts}

\usepackage{multirow}
\usepackage{array,hhline}
\usepackage{amsmath}
\usepackage{comment}
\usepackage[font={footnotesize}]{caption}
\usepackage{graphicx}
\usepackage{wrapfig}
\usepackage{floatflt}
\usepackage{soul}
\usepackage{wrapfig,lipsum,booktabs}

\usepackage{subcaption}

\usepackage{amssymb}
\usepackage{pifont}
\newcommand{\cmark}{\ding{51}}%
\newcommand{\xmark}{\ding{55}}%

\title{Self-supervised Video Object Segmentation}

\author{%
  Fangrui Zhu\thanks{Equal contribution.}\\
  Fudan University\\
  {\tt\small xiaoruirui233@gmail.com} \\
  \And
    Li Zhang$^*$ \\
    University of Oxford \\
    {\tt\small lz@robots.ox.uk}\\
    \AND
    Yanwei Fu \\
    Fudan University\\
    {\tt\small yanweifu@fudan.edu.cn}\\
    \And
    Guodong Guo \\
    West Virginia University \\
    {\tt\small guodong.guo@mail.wvu.edu}\\
    \And
    Weidi Xie\\
  University of Oxford \\
  {\tt\small weidi@robots.ox.uk}
}

\DeclareMathOperator*{\argmin}{argmin}   

\def\eg{\textit{e.g.}}

\begin{document}

\maketitle

\begin{abstract}
The objective of this paper is self-supervised representation learning, 
with the goal of solving semi-supervised video object segmentation~({\em a.k.a.}~dense tracking).
We make the following contributions:
(i)~we propose to improve the existing self-supervised approach, 
with a simple, yet more effective memory mechanism for long-term correspondence matching, 
which resolves the challenge caused by the disappearance and reappearance of objects;
(ii)~by augmenting the self-supervised approach with an online adaptation module, 
our method successfully alleviates tracker drifts caused by spatial-temporal discontinuity, 
{\em e.g.}~occlusions or dis-occlusions, fast motions;
(iii)~we explore the efficiency of self-supervised representation learning for dense tracking,
surprisingly, 
we show that a powerful tracking model can be trained with as few as 100 raw video clips~(equivalent to a duration of 11mins),
indicating that low-level statistics have already been effective for tracking tasks;
(iv)~we demonstrate state-of-the-art results among the self-supervised approaches on DAVIS-2017 and YouTube-VOS, 
as well as surpassing most of methods trained with millions of {\em manual} segmentation annotations,
further bridging the gap between self-supervised and supervised learning.
Codes are released to foster any further research~(\url{https://github.com/fangruizhu/self_sup_semiVOS}).
\end{abstract}

\section{Introduction}

Reliable and robust tracking is one of the fundamental requirements for intelligent agents,
playing a vital role in many computer vision applications, 
such as vehicle navigation, video surveillance and activity recognition. 
Generally, 
the problem is defined as to re-localize the desired object in a video sequence with the best possible accuracy, 
where the object is identified solely by its location~(bounding box) or the pixel-wise segmentation mask in the first frame,
referring as visual object tracking~(VOT)~\cite{vot} 
or semi-supervised video object segmentation~(Semi-VOS)~\cite{davis2017}, respectively. 
In this paper, 
we focus on the latter case with possibly multiple objects being present and segmented, 
and refer to it interchangeably as {\em dense tracking} from here on.

Several recent studies~\cite{eccv18_vid_color, corrflow, cvpr19_cycle_time,nips19_joint_task,cvpr20_mast} 
present promising results on self-supervised video object segmentation.
Though driven by various motivations, 
these methods can be thought of as learning pixel-wise correspondences, 
able to {\em propagate} the instance segmentation masks along the video sequence.
Leveraging the spatio-temporal coherence in natural videos,
self-supervised learning is formulated as either 
minimizing the photometric error between the raw frame and its reconstruction~\cite{eccv18_vid_color, corrflow,cvpr20_mast} 
or maximizing the cycle consistency in videos~\cite{cvpr19_cycle_time,nips19_joint_task}.
The outcome of training is an encoder to perform correspondence matching: 
the feature embedding of pixels in the “query” frame should be close to their matching pixels in ``reference'' frames
and far away from other unrelated pixels. 
Despite being simple, 
these self-supervised approaches are still challenged by the existence of spatio-temporal discontinuities, 
{\em e.g.}~occlusions, fast motion, motion blur, and textureless surfaces, 
which eventually lead accumulated errors and tracker drifts.

From this perspective, 
we hypothesize that it is desirable to augment the existing self-supervised approaches with {\em online adaptations} -- 
maintain an appearance model for the object of interest in each video sequence~\cite{Ramanan07},
continuously updating the propagated masks during inference time, {\em e.g.}~cleaning up error drifts.
To be clear, 
this is fundamentally different from the common paradigms that adapt representations acquired from large-scale supervised learning.
Here, in our case,
the appearance model is always {\em randomly initialised} for each video sequence,  
parameters of which are only trained on the imperfect masks obtained from propagation.
Thus it still falls into the self-supervised paradigm.

A general concern for the self-supervised online adaptation is 
that it could potentially overfit to the imperfect masks,
ending up with exactly the same predictions as those from mask propagations.
After all, deep networks have shown the capability of even overfitting to random labels~\cite{Zhang17Rethink}.
However, our key insight is that the model tends to learn the regularities in data faster than stochastic noises,
a phenomenon originally discovered in~\cite{cvpr18_deep_prior}.
This property is especially suitable in our self-supervised online adaptation,
as the appearance model is complementary to mask propagation, 
{\em i.e.}~less dependent on temporal coherence,
making it less likely to make the same mistakes as propagation-based models do.
We will return to this point in Section~\ref{sec:online_adaptation}.

In addition, we also show that, 
contrary to the belief that dense tracking requires good {\em semantic} representation,
we train a powerful model with as few as 100 raw videos~(with a duration of around 10mins) or even 100 still images 
through self-supervised learning, 
being able to surpass a number of approaches that are trained with in a fully supervised manner.
This is particularly remarkable, 
as it may indicate
{\bf supervised learning or semantic representation is not of the essence for dense tracking~(semi-VOS)},
helping to form a solid belief on self-supervised approaches.

To this end, we summarize our contribution as following:
(i)~we revisit the state-of-the-art self-supervised approach~(MAST~\cite{cvpr20_mast}) for dense tracking~(semi-VOS), 
and propose a simple, yet more effective memory mechanism to enable long-term correspondences,
{\em i.e.}~matching beyond pairwise frames to resolve the challenge caused by disappearance and reappearance of objects;
(ii)~propose a novel {\em self-supervised online adaptation} that complements the existing propagation-based approach,
and show it is beneficial for alleviating tracker drifts;
(iii)~conduct a pilot study to show the surprisingly high efficiency for self-supervised representation learning. 
(iv)~demonstrate state-of-the-art results among self-supervised approaches on standard benchmarks,
{\em e.g.}~DAVIS-2017 and YouTube-VOS, 
surpassing the majority of methods trained with millions of manual annotations, 
further bridging the gap between self-supervised and supervised learning.

\begin{figure}[t]
\footnotesize
\centering
\begin{subfigure}{.49\textwidth}
  \centering
   \hspace{0pt}
  \includegraphics[width=\linewidth]{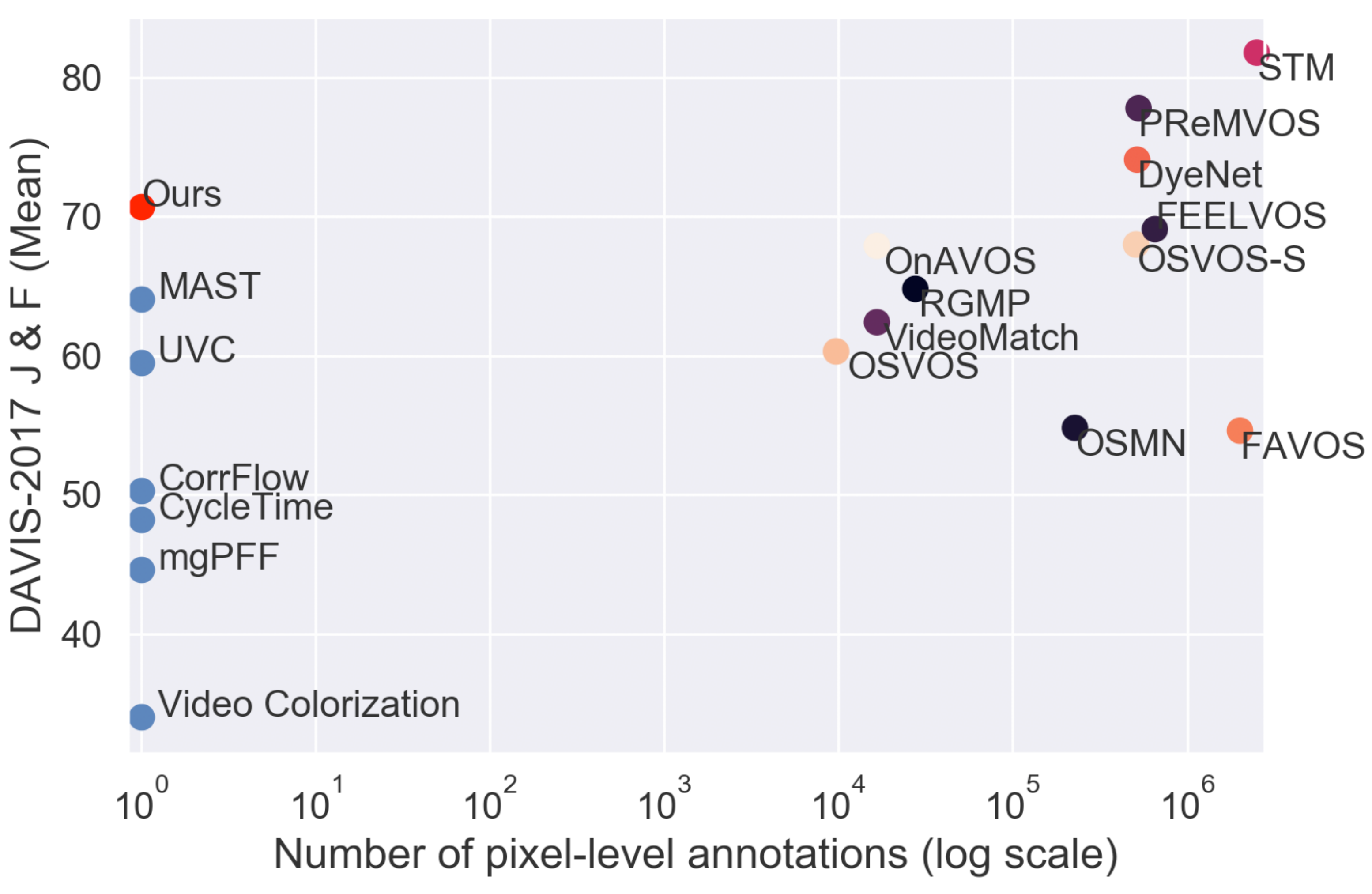}
  \caption{Comparison on DAVIS-2017.}
  \label{fig:teaser_1}
\end{subfigure}%
\hfill
\begin{subfigure}{.5\textwidth}
  \centering
  \includegraphics[width=\linewidth]{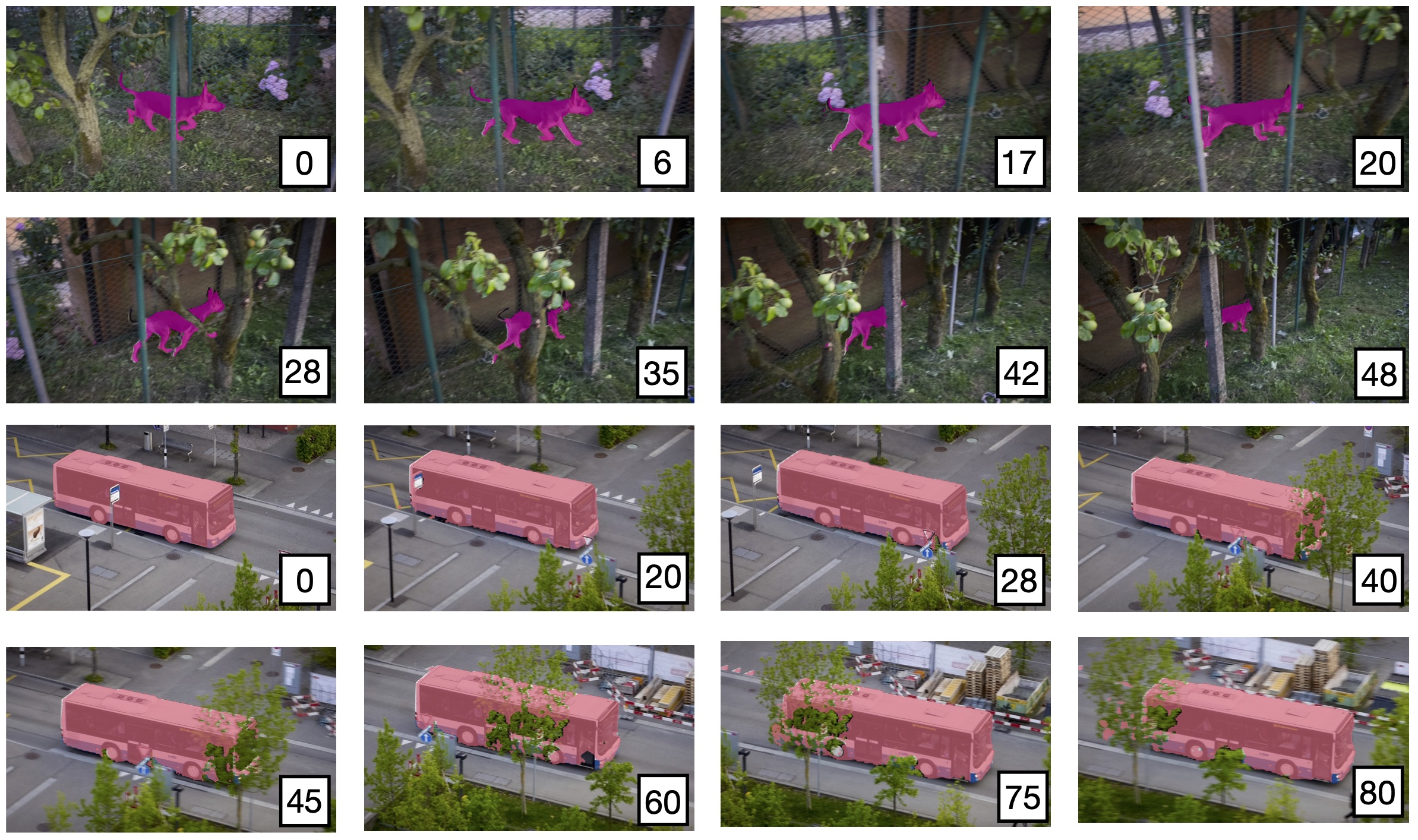}
  \vspace{-4pt}
  \caption{Qualitative results.}
  \label{fig:teaser_2}
\end{subfigure}
\vspace{-2pt}
\caption{
{\bf Notations}:~OnAVOS~\cite{bmvc17_OnAVOS}, 
OSVOS~\cite{cvpr17_OSVOS}, 
FAVOS~\cite{Cheng18},
CINM~\cite{Bao18}, 
VOSwL~\cite{Khoreva18},
DyeNet~\cite{Li18ECCV}, 
PReMVOS~\cite{Luiten18},
OSVOS-S~\cite{Maninis18}, 
OSMN~\cite{Yang18},
RGMP~\cite{Oh18},
Video Colourization~\cite{eccv18_vid_color},
AGAME~\cite{Johnander19},
mgPFF~\cite{Song19}, 
CorrFlow~\cite{corrflow}, 
RVOS~\cite{Ventura19},
FEELVOS~\cite{cvpr19_feelvos}, 
SiamMask~\cite{cvpr19_siammask},
CycleTime~\cite{cvpr19_cycle_time}, 
RANet~\cite{Wang19RA},
UVC~\cite{nips19_joint_task},
MAST~\cite{cvpr20_mast}.}
\label{fig:teaser}
\vspace{-15pt}
\end{figure}

\section{Related work}
\vspace{-5pt}
\noindent
\textbf{Self-supervised representation learning } 
has recently shown to be an promising alternative to supervised learning on a variety of downstream tasks~\cite{Oord18,Henaff19,moco,Chen20}.
In the literature, 
numerous pretext tasks have been proposed for learning semantics from free supervision signals residing in images and videos. 
These include creating pseudo classification labels~\cite{iclr18_rotation}, 
exploring spatial context in images~\cite{iccv15_contextpred} 
and temporal ordering in videos~\cite{eccv16_jigsaw}, 
such as predicting jigsaw puzzles~\cite{aaai19_puzzle}, shuffled frames~\cite{eccv16_shuffle}, motion~\cite{cvpr19_motion}, 
the arrow of time~\cite{cvpr18_arrowtime}, predicting future representations~\cite{Vondrick16b,dpc}.

\noindent
\textbf{Semi-supervised video object segmentation (Semi-VOS) }
aims to re-localize one or multiple targets that have been specified in the first frame of a video with pixel-wise segmentation masks.
Prior works can be roughly divided into two categories,
one is based on mask propagation~\cite{iccv19_stm,cvpr19_feelvos,cvpr19_siammask},
and the other is related with few shot learning or online adaptations~\cite{tpami18_osvos-s,bmvc17_OnAVOS,cvpr17_OSVOS}.
Commonly, 
extensive human annotations are required to train such systems.
More specifically, 
they generally adopt a deep neural network with its backbone network ({\em e.g.}~ResNet) pretrained on ImageNet~\cite{russakovsky2015imagenet} 
and finetune the whole framework on COCO~\cite{lin2014microsoft}, 
DAVIS~\cite{davis2017} and Youtube-VOS~\cite{eccv18_s2s}, {\em etc}.
Alternatively, 
recent approaches~\cite{eccv18_vid_color,cvpr19_cycle_time,nips19_joint_task,cvpr20_mast} 
based on self-supervised learning have shown great potentials.

\noindent
\textbf{Learning with imperfect segmentation }
has been studied for aerial images~\cite{Mnih12, Chen19} in the literature.
Here, we share the same spirit that the proposed self-supervised online adaptation module 
can only be trained on imperfect masks generated from propagation through correspondence matching.

\section{Method}
In this section, 
we first review the previous self-supervised approaches for dense tracking,
specifically, the ones target on learning pixel-wise correspondences through frame reconstruction~(Section~\ref{sec:background}).
Next, in Section~\ref{sec:memory},
we propose a simple, momentum memory mechanism that enables long-term correspondence matching, 
yet without incurring the bottleneck from the physical hardware memories.
Finally, in Section~\ref{sec:online_adaptation},
we describe the proposed self-supervised online adaptation module.

\subsection{Learning correspondence through reconstruction}
\label{sec:background}

In the recent work~\cite{eccv18_vid_color, corrflow, cvpr20_mast},
learning pixel-wise correspondence in videos has been formulated as the outcome of frame reconstruction.
Specifically, with the use of certain information bottleneck, 
each pixel from the `query' frame is forced to find pixels that can best reconstruct itself in one or multiple `reference' frames.

Mathematically, 
given a pair of frames from a video clip, {\em e.g.}~$\left \{ I_{t-1}, I_{t} \right \} \in \mathcal{R}^{H \times W \times 3}$,
we parameterise the feature encoder with a ConvNet~($\Phi(\cdot;\theta)$), 
{\em i.e.}~$f_{t} = \Phi \left ( g\left ( I_{t} \right );\theta  \right)$, where $f_t \in \mathcal{R}^{h \times w \times d}$~(height, width and channels respectively), 
$g\left ( \cdot \right )$ denotes a bottleneck that prevents information leakage.
For instance, in~\cite{eccv18_vid_color}, it refers to the RGB2Gray operation;
in~\cite{cvpr20_mast}, 
a simple channel-wise dropout in Lab colour space has shown to be surprisingly powerful.

To this end, an affinity matrix~($A$) is computed as a soft attention, 
denoting the strength of the similarity between pixels in `query' frame~($I_t$) and those in the `reference' frame~($I_{t-1}$).
Leveraging the spatio-temporal coherence in videos, 
a pixel $i$ in $I_t$ can be represented as a weighted sum of pixels in $I_{t-1}$.
To avoid abuse of notations, 
spatial positions in the frame are represented with a single character~({\em e.g.}~$i, j$).
Normally, we only consider pixels within a spatial neighbourhood to $i$, 
{\em i.e.}~$\mathcal{N} = \{\forall n \in I_{t-1}, |n - i| < c\}$:
\begin{align}
    \hat{I}_{t}^{i}&=\sum _{j \in \mathcal{N}} A_{t}^{ij}I_{t-1}^{j}\\
    A_{t}^{ij}&=\frac{\textup{exp}\left \langle f_{t}^{i},~f_{t-1}^{j} \right \rangle}{\sum _{n \in \mathcal{N}} \textup{exp} \left \langle f_{t}^{i},~f_{t-1}^{n} \right \rangle}  
\end{align}
where $c$ refers to the radius of neighborhood, and $A_t \in \mathcal{R}^{hw \times 4c^2}$.
The outcome of training is an encoder~($\Phi(\cdot; \theta)$) to perform correspondence matching by minimizing some photometric loss~({\em e.g.}~Huber loss), 
{\em i,e.}~the feature embedding of pixels in the “query” frame should be close to their matching pixels in ``reference'' frames
and far away from other unrelated pixels. 
\begin{align}
\hat{\theta} = \argmin_\theta \mathcal{L} (I_t, ~\hat{I}_t)
\end{align}
During inference, the same affinity matrix is computed to {\em propagate} the instance segmentation mask:
\begin{align}
    \hat{y}_{t}^{i}=\sum _{j \in \mathcal{N}} A_{t}^{ij}y_{t-1}^{j}
    \label{eq:propagation}
\end{align}
$y_t$ refers to the segmentation mask for the $t$-th frame.

\begin{figure}[t]
    \begin{center}
        \includegraphics[width=.85\textwidth]{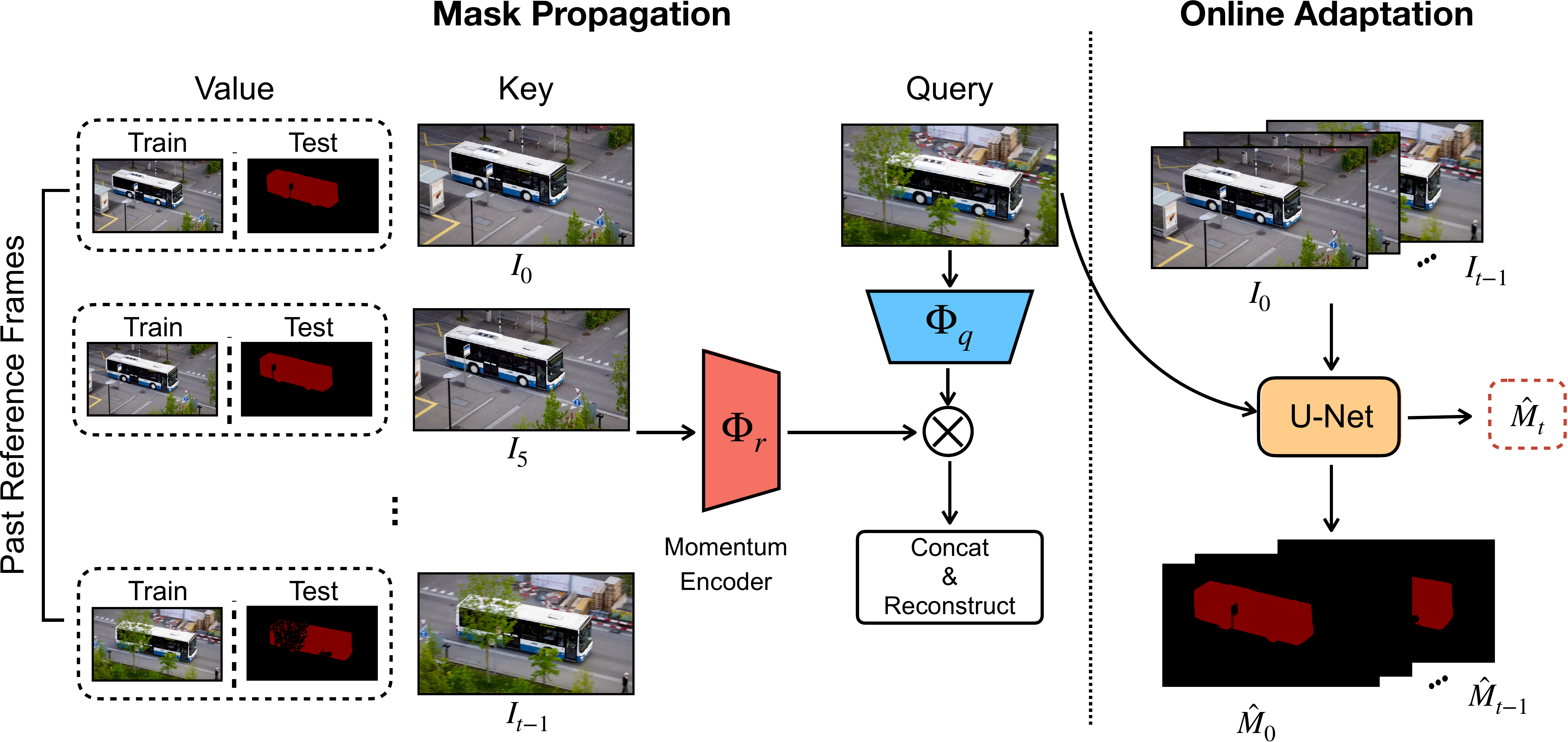}
    \end{center}
     \vspace{-5pt}
    \caption{Schematic illustration of the proposed method, 
    {\em i.e.}~self-supervised representation learning and mask propagation,
    and online adaptation during inference time.}
    \vspace{-10pt}
\end{figure}

\subsection{Occlusion-aware mask propagation}
\label{sec:memory}
One issue with learning correspondence from pairwise frames is that, 
it can not effectively deal with object disappearance and reappearance.
For example, if the object is occluded in one frame~($I_t$), and re-appear in the next one~($I_{t+1}$),
pairwise matching is deemed to fail as the object in $I_{t+1}$ cannot find its counter part in previous frame~($I_t$).
In psychology, this refers to a fundamental concept -- object permanence,
the capability of understanding that objects continue to exist even when they cannot be seen, 
heard, touched, smelled or sensed.

Computationally, a straightforward idea is to maintain an external memory,
caching multiple frames for potential correspondence matching.
In this paper, 
we propose a momentum memory, 
which significantly simplify the memory training in~\cite{cvpr20_mast}, 
able to cache as long as the entire video sequence, 
without incurring the bottleneck from the limitation of physical GPU memory.

Formally,  
given a query frame $I_q$ and an external memory bank with $K$ frames, 
{\em e.g.}~$I_r = \left \{I_{1}, \dots ,I_K \right \} \in \mathcal{R}^{K \times H \times W \times 3}$,
we parameterize two ConvNets, {\em e.g.}~$\Phi_q(\cdot;\theta_q)$ and $\Phi_r(\cdot; \theta_r)$, 
to compute representations for query and reference frames:
\begin{equation}
f_q, f_r = \Phi_q(I_q; ~\theta_q), \Phi_r(I_r; ~\theta_r)
\end{equation}
The reconstruction of the query frame $I_q$ in pixel $i$ becomes:
\begin{align}
    \hat{I}_{q}^{i} &=\sum _{j \in \mathcal{N}} A_{q}^{ij}I_{r}^{j} \\
    A_{q}^{ij} &=\frac{\textup{exp}\left \langle f_{q}^{i},~f_{r}^{j} \right \rangle}{\sum _{n \in \mathcal{N}} \textup{exp} \left \langle f_{t}^{i},~f_{r}^{n} \right \rangle}
\end{align}
where $A_q \in \mathcal{R}^{hw \times K4c^2}$, $\mathcal{N} = \{\forall n \in  I_r, |n - i| < c \}$.
In practice, 
the number of reference frame can be varied, 
we use $5$~(with the first frame always included) 
as it provides a balance between performance and computation~(shown in Section~\ref{sec:ablation}).

While training, $\theta_q$ is updated with standard back-propagation, 
and $\theta_r$ is updated with momentum, similar to~\cite{moco}:
\begin{equation}
    \theta _{r} \leftarrow m\theta _{r} + \left ( 1-m \right )\theta _{q}
\end{equation}
Here $m\in \left [ 0,1 \right )$ is a momentum coefficient.
In our case, we use $m=0.999$ in the experiments~(ablation studies have been shown in Section~\ref{sec:ablation}). 
To this end, we can train a complete model,
the feature encoder~($\Phi_q(\cdot; \theta_q), \Phi_r(\cdot; \theta_r)$) by self-supervised frame reconstruction,
and use it to {\em propagate} the mask from initial frame to subsequent ones in the video sequence.

\subsection{Self-supervisd online adaptation}
\label{sec:online_adaptation}
Despite being effective for understanding object permanence, 
the memory mechanism still suffers from spatio-temporal discontinuity, 
{\em e.g.}~occlusion or disocclusions~(regions that are originally occluded in all previous frames become visible).
Under these scenarios, 
self-supervised approaches based on low-level statistics~(photometric consistency in frame reconstruction) 
become insufficient to establish reliable correspondences,
eventually leading to accumulated errors and tracker drifts.
As an extension to the propagation-based tracker, 
we propose to incorporate an online adaptation module to fit the appearance of the target instances~\cite{Ramanan07}.

Specifically,
during the inference stage, 
for frame $I_t$ in a test video sequence, 
we can easily obtain the mask predictions for all previous frames from correspondence propagation~(Eq~\ref{eq:propagation}),
denoted as $\{(I_1, y_1), (I_2, \hat{y}_2), \dots, (I_{t}, \hat{y}_{t})\}$,
where the $y_i, \hat{y}_i$ refer to the ground-truth segmentation mask and prediction from correspondence matching~(mask propagation) respectively.
Note that, the ground-truth segmentation mask is only available for the first frame in the video sequence.

With these segmentation masks, 
we are able to train an appearance model~({\em e.g.}~frame-wise segmentation model, $\Psi(\cdot; \theta_o)$) 
from \emph{scratch} for each video sequence, where:
\begin{align}
 \hat{\theta}_o = \argmin_{\theta_o} \mathbb{E} \left[ \mathcal{L} (\Psi(I_i; \theta_o), ~\hat{y}_i )\right]
\end{align}
In practice, 
we adopt a simple U-Net~\cite{unet} as the appearance model~(details in Section~\ref{sec:setup}).
Note that, this is fundamentally different from the the previous online adaptation approaches~\cite{bmvc17_OnAVOS},
as we do not require any supervised pre-training or complex sample minings.
In fact, for each video sequence, 
there is always one tailored appearance network training from \emph{scratch}.

\par{\bf{Discussion. }}
A general concern for such self-supervised online adaptation is that it could potentially overfit to the imperfect masks,
ending up with same predictions as the mask propagation.
However, in practice, 
as the online adaptation module is complementary to the propagation-based approach,
we observe it is indeed possible to train a deep network on single video sequence with imperfect masks.
To be specific,
propagation can usually solve smooth object deformations with its built-in spatio-temporal coherence,
but fails when faced with discontinuity~(regions that have been invisible till this time point).
In contrast, the appearance model treats each frame independently,
as the same object instances in a video sequence often show far higher similarity~(self-similarity) than with the background, 
the built-in prior in ConvNets therefore tends to grasp the regularities in data before overfitting to noises,
a phenomenon originally observed in~\cite{cvpr18_deep_prior}.
We experimentally show this phenomenon in Section~\ref{sec:online}.

\section{Experiments}
In this section,  we first introduce the datasets and implementation details in Section~\ref{sec:setup}, 
and then describe the training process for learning dense correspondence and the online adaptation in Section~\ref{sec:online}.
We further examine the effects of different components in Section~\ref{sec:ablation}.
With the optimal settings,
we conduct a pilot study with the goal of better understanding the essence of tracking~(Section~\ref{sec:essence}),
specifically, on the role of semantic supervision.
Lastly,
we show comparison to the state-of-the-art Semi-VOS approaches in Section~\ref{sec:sota}.

\subsection{Experimental setup}
\label{sec:setup}

\noindent
{\bf Datasets and evaluation. }
We conduct experiments on two widely-used datasets, 
DAVIS-2017~\cite{davis2017} and YouTube-VOS~\cite{eccv18_s2s}.
Specifically, 
DAIVS-2017 contains 150 HD videos with over 30K instance segmentation annotations,
and YouTube-VOS has over 4,000 HD videos of 90 semantic categories with over 190K instance segmentation annotations. 
In this paper, 
all training has been done on the YouTube-VOS training set in a self-supervised manner, 
{\em i.e.}~{\em zero} groundtruth segmentation annotation is used during training time.

For evaluation, 
we benchmark on the official semi-supervised video object segmentation protocol of DAVIS-2017, 
and YouTube-VOS~2018 val set,
that is, 
ground-truth segmentation mask is only given in the first frame, 
and the goal is to predict the object mask in subsequent frames. 
Standard evaluation metrics are used, 
namely, region similarity~($\mathcal{J}$) and contour accuracy~($\mathcal{F}$). 

\noindent
{\bf Implementation details. } 
For learning correspondence, 
we adopt a ResNet-18 as our feature encoder, same as in~\cite{cvpr20_mast},
it produces feature embeddings with spatial resolution of $1/4$ of the original image. 
As pre-processing,  
we randomly crop the original images and resize them to the size of $384 \times 384$.
The same data augmentation is applied as in~\cite{cvpr20_mast}, 
\eg ~$Lab$ space inputs for both the query and reference frames with random colour jittering~($p=0.3$) and channel dropout~($p=0.5$).

In detail,
we first pretrain the feature encoder with pairwise inputs for 120K iterations and batch size of 48 with Adam optimiser.
After pretraining, we finetune the model with a momentum memory, 
where multiple previous frames are treated as references, 
and a smaller learning rate of $1 \times 10^{-4}$ is applied for another 10K iterations. 
The window size for neighbourhood is set to 25pix in our training.

During inference time,
the affinity matrix is computed between the query and reference frames in the memory, 
and further used for propagating the desired pixel-level segmentation masks. 
At the same time, 
we start the online adaptation procedure, 
{\em i.e.}~train an appearance model on each video sequence with the frames in memory as inputs 
and their predicted instance masks as the pseudo ground-truth.
For simplicity, the appearance model uses a U-Net~\cite{unet} with ResNet-18 backbone.
All inputs are resized to $480 \times 480$ with random colour jittering, 
gray-scaling and horizontal flipping.
Note that, we train the appearance model {\em from scratch} for each test sequence, 
without updating the feature encoder~($\Phi(\cdot)$).
Adam optimiser is used, with an initial learning rate of $2\times 10^{-4}$, decaying with the step size of 50. 
The entire adaptation process lasts for around 200 iterations, 
and the outputs from appearance model are treated as final predictions.

\subsection{Self-supervised online adaptation}
\label{sec:online}
\begin{figure}[!htb]
\footnotesize
\centering
\vspace{-6pt}
\begin{subfigure}[t]{.36\textwidth}
  \centering
  \hspace{-15pt}
  \includegraphics[width=\linewidth]{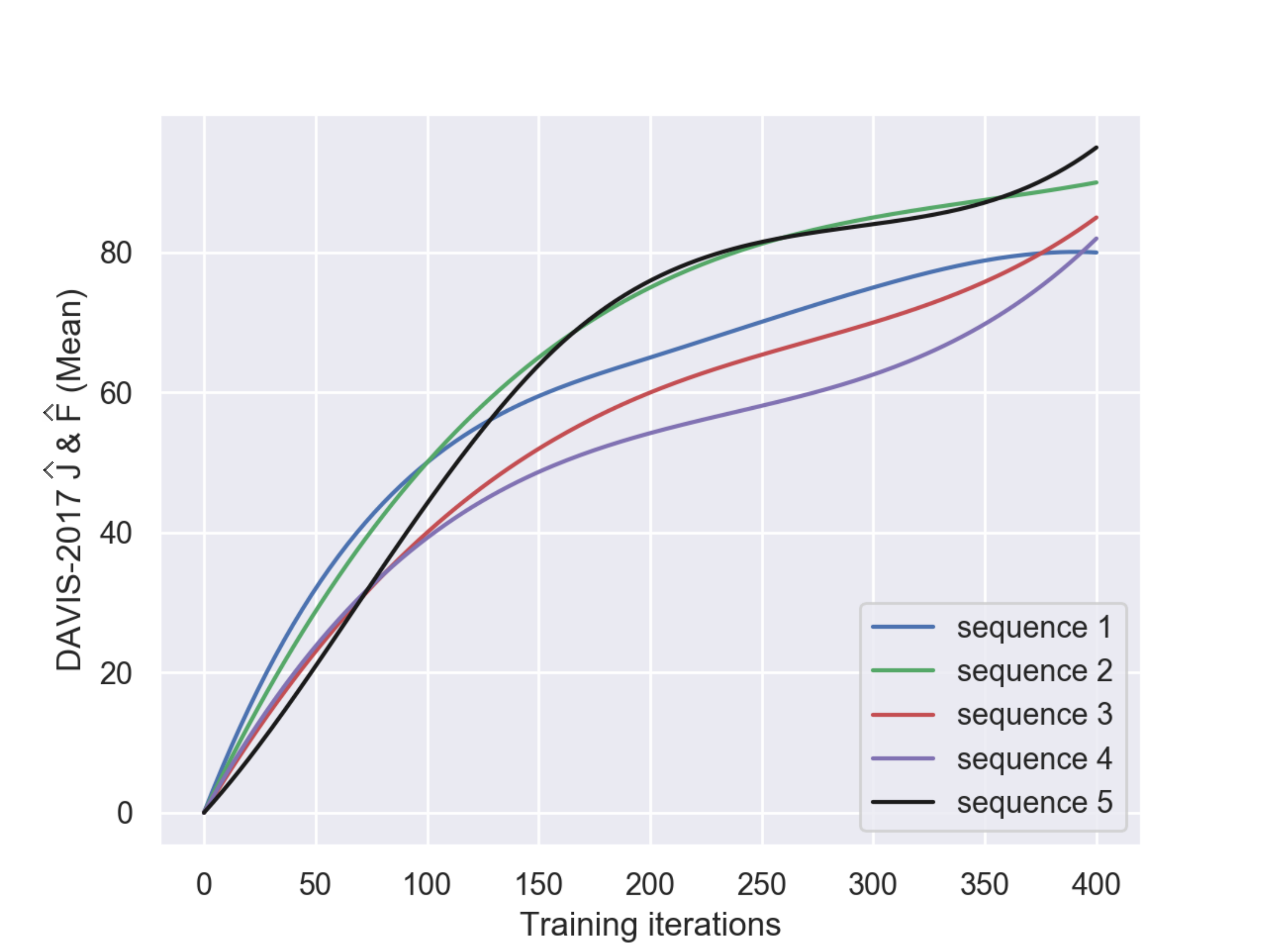}
  \vspace{-5pt}
  \caption{}
  \label{fig:online_curve_1}
\end{subfigure}%
\hspace{15pt}
\begin{subfigure}[t]{.36\textwidth}
  \centering
    \hspace{-15pt}
  \includegraphics[width=\linewidth]{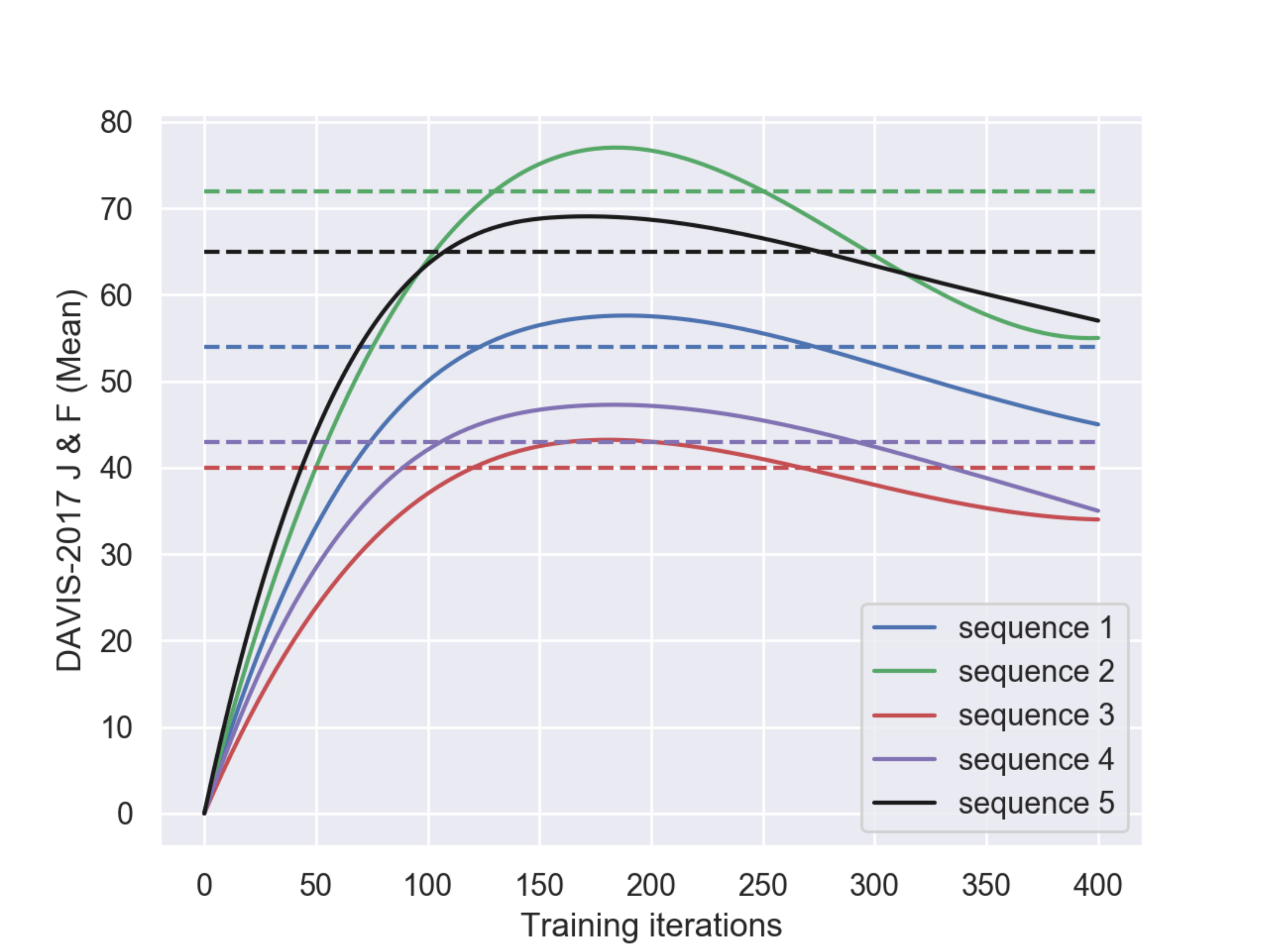}
    \vspace{-5pt}
  \caption{}
  \label{fig:online_curve_2}
\end{subfigure}
\vspace{-6pt}
\caption{Learning process of the self-supervised online adaptation module.
In (a), $\hat{\mathcal{J}}$\&$\hat{\mathcal{F}}$~(mean) is computed between the predictions 
  from appearance model and the propagated masks~(pseudo goundtruth).
In (b), $\mathcal{J}$\&$\mathcal{F}$~(mean) is computed between the predictions 
  from appearance model and manual ground-truth masks acquired from an {\em oracle}.
  Dash lines denote the baseline results between propagated masks and groundtruth.}
\label{fig:online_curve}
\end{figure}

In this section, 
we aim to validate the idea of training an appearance model from imperfect segmentation masks.
Specifically, 
we assume there exists an {\em oracle} with the access to ground-truth, 
such that the learning process for online adaptation can be monitored.
In Figure~\ref{fig:online_curve}, 
we compare the prediction from online adaptation~(appearance model) with mask propagation~(Figure~\ref{fig:online_curve_1}),
and with ground-truth masks provided by the {\em oracle}~(Figure~\ref{fig:online_curve_2}),
dash lines can be seen as the baseline, indicating the raw performance of mask propagation.
Note that, manual segmentation masks are only used to plot this curve, 
but {\em never} be used for training the appearance model.

As can be seen in Figure~\ref{fig:online_curve_2}, 
the appearance model quickly absorbs information from the propagated segmentation masks,
outperforming the baselines~(shown by dash lines) from mask propagation.
However, with more training iterations, the model starts to give inferior predictions.
This phenomenon confirms our conjecture that the structure of ConvNets has enforced a strong prior on image statistics~\cite{cvpr18_deep_prior}, 
which enables the model to grasp the regularities~(self-similarity of the instances) in data before overfitting to error drifts.
Another phenomenon is, 
the best appearance models can always be obtained at the stage, 
where the learning process starts to get slow, roughly around $150$-$250$ iterations.
In the following online adaptation experiments, we thus decide to train only 200 iterations.

\vspace{-10pt}
\subsection{Ablation studies}
\label{sec:ablation}
To examine the effectiveness of different components, 
we conduct a series of experiments by adding one component at a time,
\eg~momentum memory, online adaptation. 
All models are trained on YouTube-VOS, and evaluated on DAVIS-2017 semi-supervised video object segmentation protocol.

\setlength{\tabcolsep}{1pt}
\begin{table}[!htb]
\footnotesize
     \begin{minipage}{.48\linewidth}
     \centering
     \vspace{-4pt}
    \begin{tabular}{lccc}
    \toprule
    \multirow{2}{*}{Variants} & \multicolumn{3}{c}{DAVIS-2017} \\
     \cmidrule(lr){2-4} 
                   & $\mathcal{J} \& \mathcal{F}$ &  $\mathcal{J}$(Mean) & $\mathcal{F}$(Mean) \\
    \midrule
    MAST pairwise~\cite{cvpr20_mast}  & 59.6&  57.3  &  61.8  \\
    +~memory~(MAST~\cite{cvpr20_mast}) &   63.8 &  61.2  &   66.3 \\
    \midrule
 Ours~(pairwise)    & 60.4 & 58.7  &  62.0  \\
+~momentum memory &   68.2 &66.5  &   69.9 \\
+~online adaptation &  70.7& 68.2 & 73.1 \\
    \bottomrule
    \end{tabular}
    \vspace{6pt}
     \caption{Effects of the proposed modules, {\em e.g.}~momentum memory, online adaptations.}
     \label{tab:ablation_1}
    \end{minipage}
    \hspace{10pt}
    \begin{minipage}{.47\linewidth}
      \centering
	\begin{tabular}{lcccc}
	\toprule 
	\multirow{2}{*}{variants} & \multirow{2}{*}{number} & \multicolumn{3}{c}{DAVIS-2017} \\
	\cmidrule(lr){3-5} 
	& 	   &  $\mathcal{J} \& \mathcal{F}$  & $\mathcal{J}$(Mean)      & $\mathcal{F}$(Mean)    \\
  	  \midrule
  	   \multirow{3}{*}{\# ref frames} & 1    &   60.4 & 58.7  &   62.0       \\
  	   						  & 3   &  65.0 &   63.6  &   66.2       \\
                         			          & 5      &  68.2 &   66.5   & 69.9  \\
                          			         & t-1    &   66.4 &  65.0   &  67.8             \\             
    	\bottomrule
	\end{tabular}
	\vspace{6pt}
        \caption{Effects of the number of reference frames in memory while propagating masks to $t$-th frame.
        $t-1$ refers to the extreme case of caching all previous frames.}
        \label{tab:mem}
    \end{minipage}%
     \caption*{}
     \vspace{-25pt}
\end{table}

\begin{figure*}[!htb]
\footnotesize
\begin{center}
\begin{tabular}{c}
\includegraphics[width=\textwidth, height=0.1\textwidth]{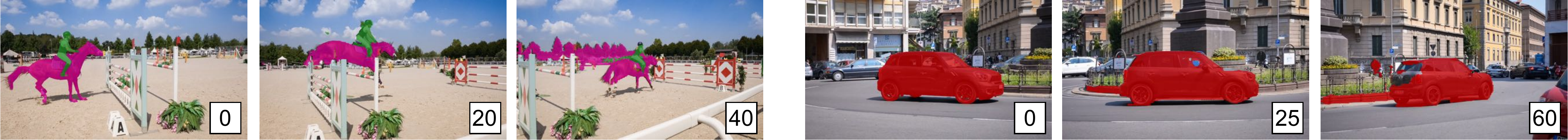} \\
\multicolumn{1}{c}{\footnotesize{(a) Qualitative results from pairwise model}} \\
\\ [-5pt]
\includegraphics[width=\textwidth, height=0.1\textwidth]{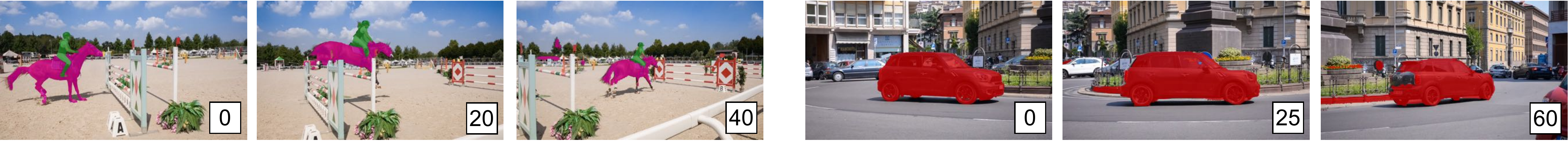} \\
\multicolumn{1}{c}{\footnotesize{(b) Qualitative results by adding momentum memory}} \\
\\ [-5pt]
\includegraphics[width=\textwidth, height=0.1\textwidth]{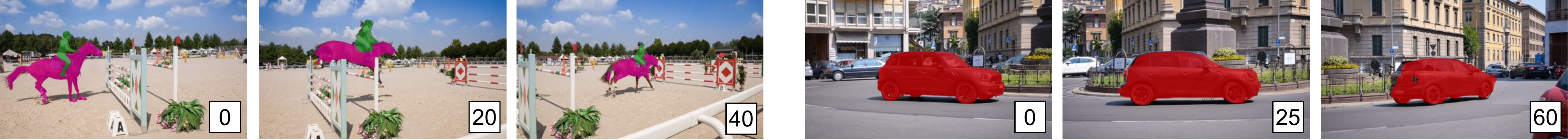} \\
\multicolumn{1}{c}{\footnotesize{(c) Qualitative results by introducing online adaptation}} \\
\end{tabular}
\vspace{-.2cm}
\caption[]{Qualitative comparison for the effectiveness of the proposed modules.}
\vspace{-.8cm}
\label{fig:com_vis}
\end{center}
\end{figure*}

We first re-implement the pairwise model from MAST~\cite{cvpr20_mast}~(59.6 vs 60.4).
As shown in Table~\ref{tab:ablation_1},  
the proposed momentum memory brings a significant performance boost, from 60.4 to 68.2 on $\mathcal{J} \& \mathcal{F}$.
In addition, 
the proposed self-supervised online adaptation module further improves on the high baseline~(from 68.2 to 70.7 on $\mathcal{J} \& \mathcal{F}$).

Next, we study the effect of using different number of reference frames in the memory bank.
As shown in Table~\ref{tab:mem}, 
it is clear that the memory module~(with more than one reference frame) plays a vital role for this task, 
but the model is fairly robust to the number of reference frames. 
Thus, 
we choose to use $5$ reference frames in the following experiment, as it gives the optimal performance, 
yet still be efficient during inference time. 

In Figure~\ref{fig:com_vis}, we show the qualitative comparisons by adding one components at a time.
The model based on pairwise frame propagation suffers serious error drifts, 
{\em e.g.}~horse masks drift to the background, the car masks drift to the road.
For both cases, the errors are caused by the dis-occlusion, 
{\em i.e.}~the regions that are originally hidden start to show up.
With the help of momentum memory and online adaptation,  
errors have been significantly alleviated.

\subsection{Data efficiency for self-supervised training}
\label{sec:essence}

\begin{wraptable}{r}{6.0cm}
\footnotesize
\vspace{-10pt}
\begin{tabular}{lccc}
\toprule
\multirow{2}{*}{Training} & 
\multirow{2}{*}{Variants} &
\multicolumn{2}{c}{DAVIS-2017} \\
 \cmidrule(lr){3-4} 
 &  & $\mathcal{J}$(Mean) & $\mathcal{F}$(Mean)  \\
    \midrule
\multicolumn{1}{c}{\multirow{2}{*}{100 videos}} & pairwise & 54.7 & 57.8 \\
\multicolumn{1}{c}{} & +~memory & 62.5 & 65.5   \\
\multicolumn{1}{c}{} & +~adaptation & 64.0 & 66.7    \\
    \midrule
\multicolumn{1}{c}{\multirow{2}{*}{100 images}} & pairwise & 53.7 & 56.2 \\
\multicolumn{1}{c}{} & +~memory &  60.0 &  63.1 \\
\multicolumn{1}{c}{} & +~adaptation & 62.1 & 64.5    \\
\hline
\end{tabular}
\caption{Efficient self-supervised learning, evaluated on DAVIS-2017 validation sets.}
\label{tab:ablation_data}
\vspace{-10pt}
\end{wraptable} 

In this section, 
we conduct a study for learning powerful representations under a low-data regime.
Specifically,
we consider experiments in two settings, 
namely, only training a small number of images or real videos. 
For images, 
we use simple homography transformations to augment images into video sequences,
and the proposed model is then trained on these simulated sequences. 
In comparison, we also train the same model with only 100 raw videos~(a duration of around 10mins).

As shown in Table~\ref{tab:ablation_data},
despite only a small number of images or videos are used for training,
the model still performs remarkably well, 
{\em i.e.}~matching the previous state-of-the-art self-supervised learning approach~\cite{cvpr20_mast}, 
and outperforming many supervised methods in Table~\ref{tab:quan_result_davis2017}.
It implies that supervised learning or semantic representation may be not essential in dense tracking,
since there is merely any semantic information under such a low-data training regime.

\vspace{-5pt}
\subsection{Compare with state-of-the-art}
\label{sec:sota}
In this section, 
we compare with state-of-the-art approaches on DAVIS-2017~(Table~\ref{tab:quan_result_davis2017}) and YouTube-VOS~(Table~\ref{tab:ytb}) semi-VOS benchmarks.
Note that, 
there has been a rapid progress in this research line, 
here, we only try to compare with the recent state-of-the-art approaches.
As different architectures and training set were used in the previous work, 
which makes fair comparison extremely difficult.

\setlength{\tabcolsep}{2pt}
\begin{table}[!htb]
\footnotesize
  \centering
  \begin{tabular}{cccccccccc}
  \toprule
   Method  & Date &  Arch. & Sup. & Dataset & $\mathcal{J} \& \mathcal{F} $ & $\mathcal{J}$(Mean) & $\mathcal{J}$(Recall) & $\mathcal{F}$(Mean) & $\mathcal{F}$(Recall) \\
   \midrule
   Vid.~Color.~\cite{eccv18_vid_color}  &2018& ResNet-18 & \xmark & K(800 hrs) & 34.0 & 34.6 & 34.1 & 32.7 & 26.8  \\
   CycleTime~\cite{cvpr19_cycle_time} &2019& ResNet-50 & \xmark& V(344 hrs) & 48.7 & 46.4 & 50.0 & 50.0 & 48.0 \\
   CorrFlow~\cite{corrflow} &2019& ResNet-18  & \xmark & O(14 hrs) & 50.3 & 48.4 & 53.2 & 52.2 & 56.0 \\
   UVC~\cite{nips19_joint_task} &2019& ResNet-18  & \xmark & K(800 hrs) & 59.5 & 57.7 & 68.3 & 61.3 & 69.8 \\
   MAST~\cite{cvpr20_mast} &2020& ResNet-18  &\xmark & Y(5.58 hrs) & 65.5 & 63.3 & 73.2 & 67.6 & 77.7 \\
  \bf Ours &2020& ResNet-18 &\xmark & DY$^*$(11 mins) & \bf 65.4  &\bf 64.0 &\bf 71.6 &\bf 66.7 &\bf 75.8\\
   \bf Ours &2020& ResNet-18 &\xmark & Y(5.58 hrs) & \bf 70.7 &\bf 68.2 &\bf 77.8 &\bf 73.1 &\bf 83.5\\
   \midrule
   ImageNet~\cite{cvpr16_resnet} &2016& ResNet-50 & \cmark & I & 49.7 & 50.3 & - & 49.0 & - \\
   OSVOS~\cite{cvpr17_OSVOS} &2017&  VGG-16  &\cmark & ID & 60.3 & 56.6 & 63.8 & 63.9 & 73.8 \\
   OnAVOS~\cite{bmvc17_OnAVOS} &2017 & ResNet-38  & \cmark & ICPD & 65.4 & 61.6 & 67.4 & 69.1 & 75.4 \\
   OSMN~\cite{cvpr18_osmn}  &2018& VGG-16 & \cmark & ICD & 54.8 & 52.5 & 60.9 & 57.1 & 66.1 \\
   OSVOS-S~\cite{tpami18_osvos-s}  &2018& VGG-16 & \cmark & IPD & 68.0 & 64.7 & 74.2 & 71.3 & 80.7 \\
   PReMVOS~\cite{Luiten18} &2018&  ResNet-101 & \cmark & ICDPM & 77.8 & 73.9 & 83.1 & 81.8 & 88.9     \\
   SiamMask~\cite{cvpr19_siammask}   &2019& ResNet-50& \cmark & ICY & 56.4 & 54.3 & 62.8 & 58.5 & 67.5 \\
   FEELVOS~\cite{cvpr19_feelvos} &2019  & Xception-65& \cmark & ICDY & 71.5 & 69.1 & 79.1 & 74.0 & 83.8 \\
   STM~\cite{iccv19_stm}   &2019 & ResNet-50 & \cmark & IDY& 81.8 & 79.2 & - & 84.3 & - \\
  \bottomrule
  \end{tabular}
  \vspace{5pt}
  \caption{Quantitative results of multi-object video object segmentation on DAVIS-2017 validation set. 
   Dataset notations: I=ImageNet,
   C=COCO, D=DAVIS, P=PASCAL-VOC, Y=YouTube-VOS, 
   DY$^*$=randomly sample 100 videos from DAVIS and YouTube-VOS, both from training sets only, 
   K=Kinetics, V=VLOG, O=OxUvA.
  Results of other methods are directly copied from~\cite{cvpr20_mast,iccv19_stm}.}
    \label{tab:quan_result_davis2017} 
      \vspace{-10pt}
\end{table}

\begin{wraptable}{r}{7.2cm}
\vspace{-15pt}
\footnotesize
\begin{tabular}[t]{cccccccccccc}
    \toprule
\multirow{2}{*}{Method} & \multirow{2}{*}{Sup.} & \multirow{2}{*}{Overall} &\multirow{2}{*}{} &\multicolumn{3}{c}{Seen} &\multirow{2}{*}{}& \multicolumn{3}{c}{Unseen} &\multirow{2}{*}{}  \\ \cmidrule{5-7} \cmidrule{9-12}
    &  &  & &$\mathcal{J}$ & & $\mathcal{F}$ & &$\mathcal{J}$ & & $\mathcal{F}$ & \\ 
    \midrule
    Vid.~Color.~\cite{eccv18_vid_color}& \xmark & 38.9 & &43.1  & &38.6 && 36.6 & &37.4 &  \\
    CorrFlow~\cite{corrflow} & \xmark & 46.6 & &50.6& & 46.6 & &43.8 & & 45.6 & \\
    MAST~\cite{cvpr20_mast} & \xmark & 64.2 & &63.9 & &64.9 & &60.3 & & 67.7 & \\
    \textbf{Ours} &  \xmark & \textbf{67.3} & &\textbf{67.2} & & \textbf{67.9} & &\textbf{63.2} & &\textbf{70.6} &\\

    \midrule
    OSMN~\cite{cvpr18_osmn} & \cmark & 51.2 & &60.0 & &60.1 & &40.6 & & 44.0 &  \\
    MSK~\cite{cvpr17_msk} & \cmark & 53.1 & & 59.9 & & 59.5 & & 45.0 & & 47.9 &\\
    RGMP~\cite{cvpr18_rgmp} & \cmark & 53.8 & & 59.5 & & - & & 45.2 & &- &  \\
    OnAVOS~\cite{bmvc17_OnAVOS} & \cmark & 55.2 & &60.1 & &62.7 & &46.6 & &51.4 &  \\
    S2S~\cite{eccv18_s2s} & \cmark & 64.4 & &71.0 & &70.0 & &55.5 & &61.2 &  \\
    A-GAME~\cite{cvpr19_agame} & \cmark & 66.1 & & 67.8 & &- & & 60.8 & & - & \\
    STM~\cite{iccv19_stm} & \cmark & 79.4 & &79.7 & &84.2 & &72.8 & &80.9 &  \\
    \bottomrule
\end{tabular}
\captionof{table}{Quantitative results on Youtube-VOS validation set.
 The proposed approach outperforms all previous self-supervised ones, 
 and compare favorably with the models trained with large-scale supervised learning.}
 \label{tab:ytb}
\vspace{-20pt}
\end{wraptable} 

As shown by the results from DAVIS-2017 and YouTube-VOS benchmarks, we can draw the following conclusions:
{\em First}, our proposed model clearly outperforms all other self-supervised methods, 
surpassing previous state-of-the-art~\cite{cvpr20_mast} by a significant margin~(as measured by $\mathcal{J} \& \mathcal{F}$, 65.5 vs 70.7 on DAVIS-2017, 64.2 vs 67.3 on YouTube-VOS). 
{\em Second}, 
compared with the supervised approaches that have been heavily trained with expensive segmentation annotations,
our proposed model trained with self-supervised learning can actually outperform the majority of them.
{\em Third},
our proposed approach is shown to be category-agnostic, 
and can generalize to unseen categories with {\em no} performance drop, as shown by testing on unseen categories~(Table~\ref{tab:ytb}).

\vspace{-5pt}
\section{Conclusion}
To summarise, in this paper, 
we have demonstrated the possibility of training competitive models for video object segmentation~(dense tracking) without using {\em any} manual annotations.
To achieve that, 
we introduced a simple, yet effective memory mechanism for long-term correspondence learning,
and the online adaptation during inference time.
In addition,
we show that, contrary to the belief that dense tracking requires good {\em semantic} representation, 
competitive models can be trained with as few as 100 raw videos,
indicating that supervised learning may not be essential for such a task.
We hope our discovery will shed light on potential research of this direction.


{\small
\bibliographystyle{unsrtnat}
\bibliography{egbib}
}

\newpage
\appendix
\section*{Appendix}
\appendix
\section{Architecture details}
\begin{figure}[!htb]
    \centering
    \includegraphics[scale=0.15]{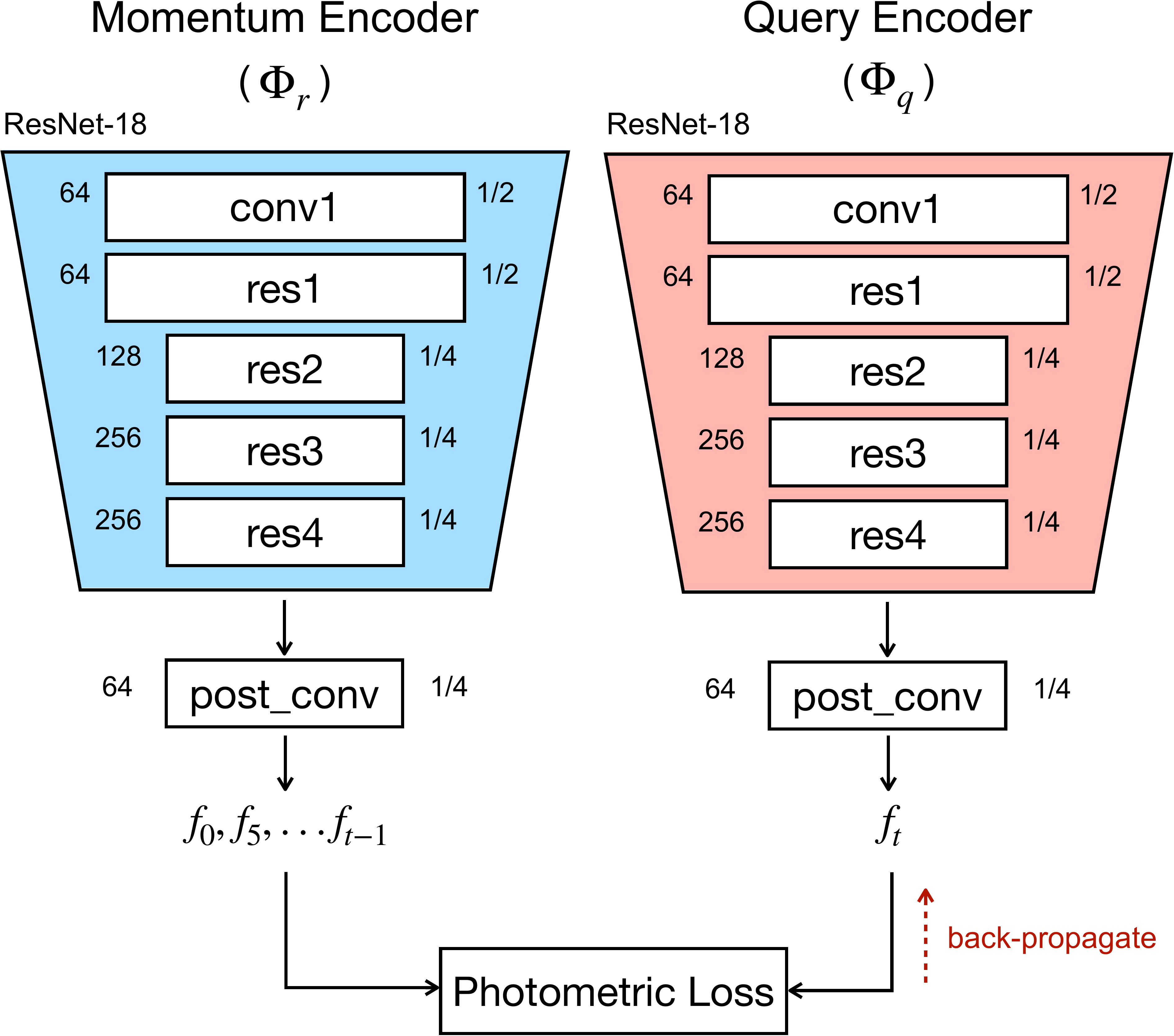}
    \caption{\label{fig:structure_enc}
    The structure of the momentum memory encoders. 
    Two trapezoids refer two ResNet-18 based encoders, 
    in which each rectangle denotes either a convolution layer or a residual block. 
    The number on the left of the rectangle shows the out channel number of that layer, {\em e.g.}~64, 128, 256, 
    while the number on the right shows the resolution ratio of the output feature map, {\em e.g.}~1/2.}
\end{figure}

We present a more detailed description of our encoder networks illustrated in Figure~\ref{fig:structure_enc}.
Specifically,
we adopt a ResNet-18 with the channel number and stride number of the residual blocks reduced in our case. 
For the appearance model, 
we take a standard U-Net~\cite{unet} structure with ResNet-18 as the backbone encoder. 

\section{More training details}

\noindent
\textbf{Occlusion-aware mask propagation.}
While training,
we use the same restricted attention mechanism as in~\cite{cvpr20_mast}, to compute the similarity between paired frames.
That is, each pixel in the current frame is only reconstructed by the pixels within a local window in the reference frames. 
In practice, we apply a $25 \times 25$ window to perform matching on $96 \times 96$ feature maps. 
Empirically, larger window size is beneficial for more accurate matching to a certain extent, 
due to the expense of more perturbation in a larger window.

\noindent
\textbf{Online adaptation.}
For each individual video sequence, 
we leverage an appearance model to perform online adaptation. 
To segment a certain video frame, 
the appearance model is trained with all the frames that have been predicted from the correspondence propagation.
The numbers of segmented objects in different video sequences are not the same, 
where we need to specify the output channel number of the appearance model for each video based on its first annotation frame.
During training, 
we use both the pixel-wise cross entropy loss and the Dice loss to optimize the model.
Under the assumption that former propagated masks contain less error accumulation and have better quality, 
we put more weights on the early frames in the video sequence, since they are less prone to drift. 
Details can be checked in the code.

\clearpage
\section{More results }
In this section, we provide more qualitative results on DAVIS-2017 and YouTube-VOS dataset.

\subsection{More qualitative results.}
\begin{figure}[!htb]
    \centering
    \includegraphics[width=1 \linewidth]{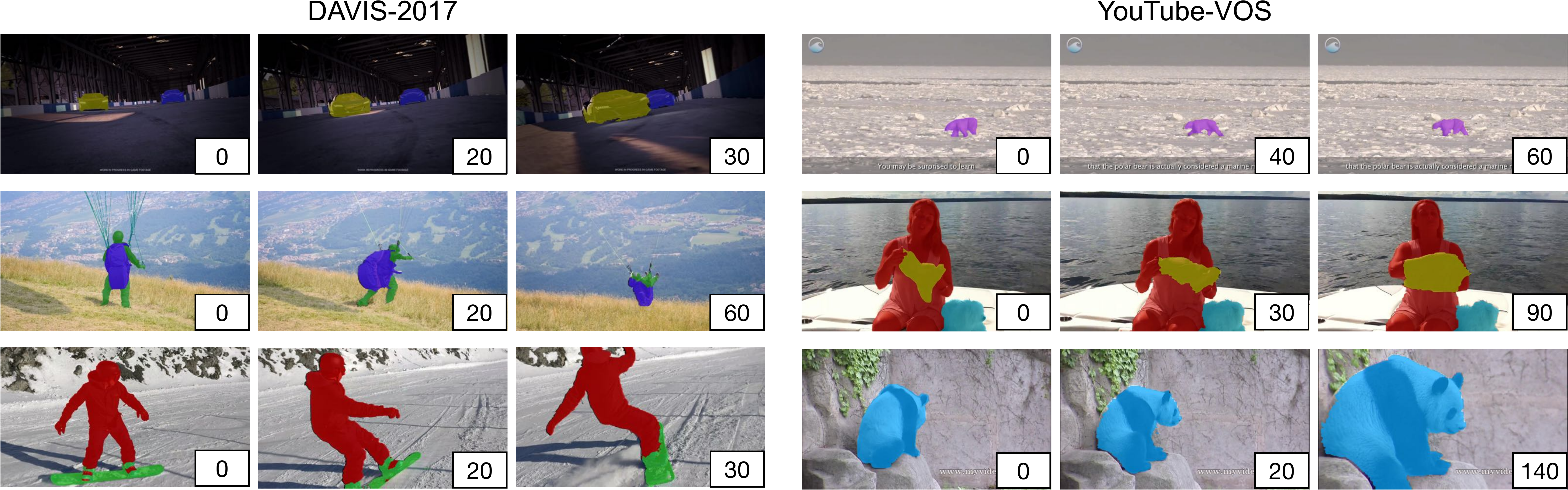}
    \caption{More qualitative results on DAVIS-2017 and YouTube-VOS datasets. 
    The first row shows the scale change and gradual occlusion case. 
    The second row shows the appearance and pose variation of the tracking object. 
    The third row shows significant camera shaking and object motion. 
    YouTube-VOS: The first row shows the motion of the white polar bear, 
    whose color is so similar with the background ice that makes it difficult for accurate tracking. 
    The second row shows the process of the person putting on her elbow guard with occlusion and dis-occlusion involved. 
    The third row shows the motion and scale change of the panda.  
    }
    \label{fig:qual_vis_both}
\end{figure}

\begin{figure}[!htb]
    \centering
    \includegraphics[width=1 \linewidth]{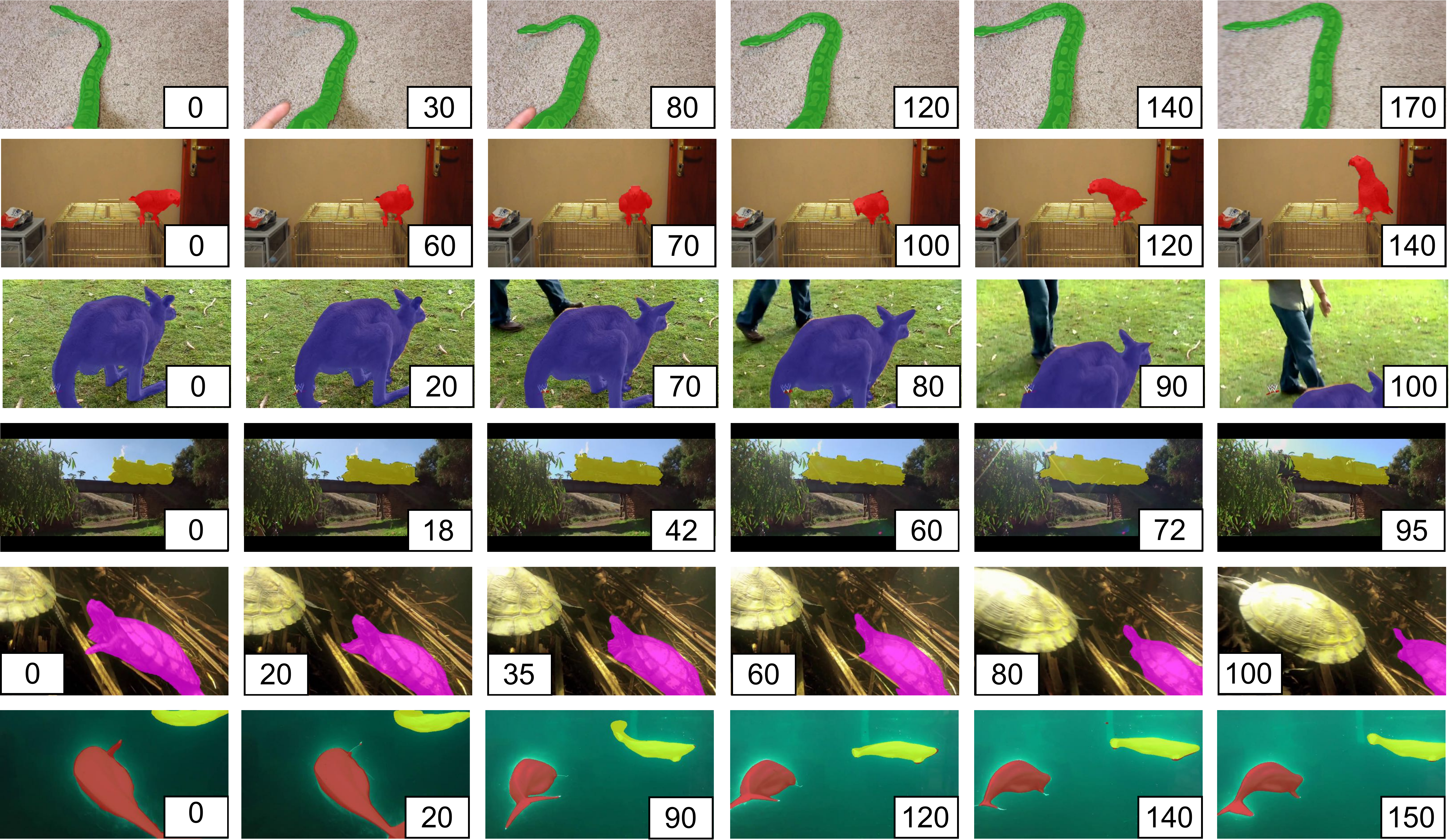}
    \caption{More qualitative results on YouTube-VOS dataset. 
    The first two rows show moving objects and pose variations among objects. 
    Row 3-5 show occlusion and dis-occlusion~(out of the scene) scenarios. 
    The sixth row shows tracking under multiple moving objects.
    }
    \label{fig:qual_vis_vos}
\end{figure}

Various difficult cases are shown in Figure~\ref{fig:qual_vis_both} and~\ref{fig:qual_vis_vos}. 
From the visualisation results, 
objects motion, large camera motion, 
cases where multiple objects co-exist, and pose variations can all be well solved by our proposed model.

\vspace{10pt}

\clearpage
\subsection{More qualitative comparison.}
\begin{figure*}[!htb]
\footnotesize
\begin{center}
\begin{tabular}{c}
\includegraphics[width=\textwidth, height=0.1\textwidth]{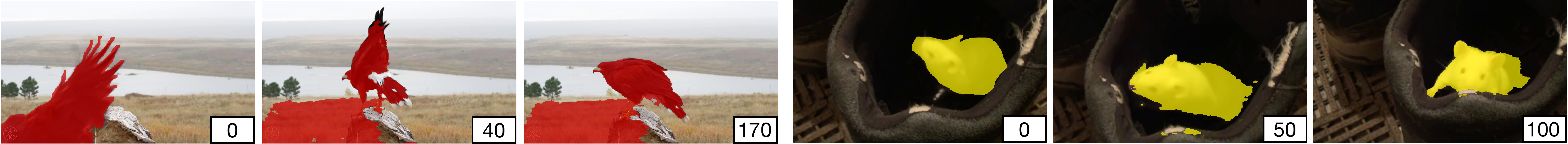} \\
\multicolumn{1}{c}{\footnotesize{(a) Qualitative results from pairwise model}} \\
\\ [-5pt]
\includegraphics[width=\textwidth, height=0.1\textwidth]{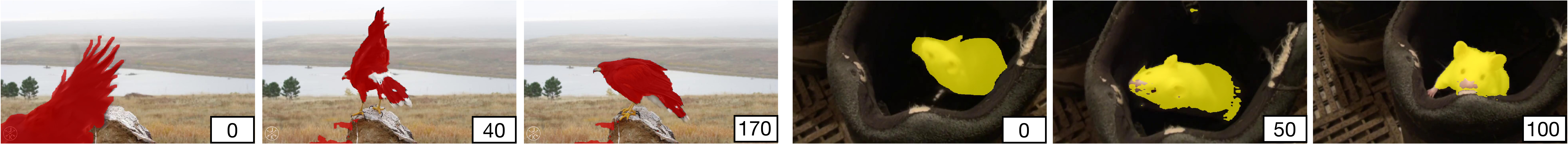} \\
\multicolumn{1}{c}{\footnotesize{(b) Qualitative results by adding momentum memory}} \\
\\ [-5pt]
\includegraphics[width=\textwidth, height=0.1\textwidth]{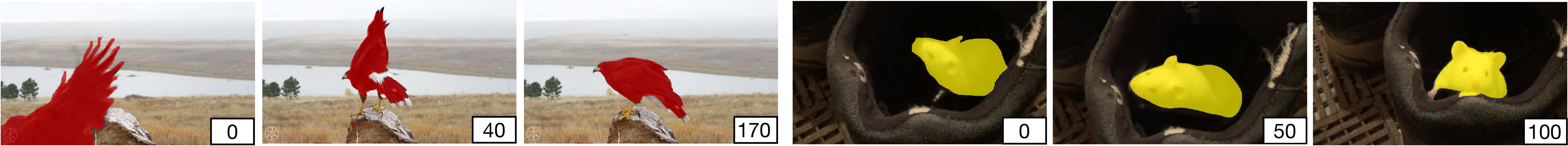} \\
\multicolumn{1}{c}{\footnotesize{(c) Qualitative results by introducing online adaptation}} \\
\end{tabular}
\caption[]{Qualitative comparison on YouTube-VOS dataset for the effectiveness of the proposed modules.}
\label{fig:com_vis_1}
\end{center}
\end{figure*}

\begin{figure*}[!htb]
\footnotesize
\begin{center}
\begin{tabular}{c}
\includegraphics[width=\textwidth, height=0.1\textwidth]{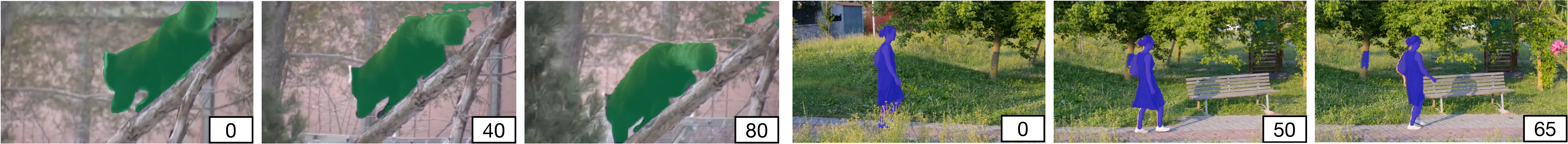} \\
\multicolumn{1}{c}{\footnotesize{(a) Qualitative results from pairwise model}} \\
\\ [-5pt]
\includegraphics[width=\textwidth, height=0.1\textwidth]{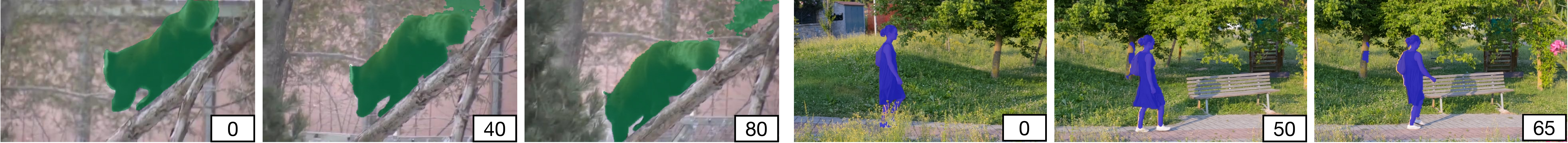} \\
\multicolumn{1}{c}{\footnotesize{(b) Qualitative results by adding momentum memory}} \\
\\ [-5pt]
\includegraphics[width=\textwidth, height=0.1\textwidth]{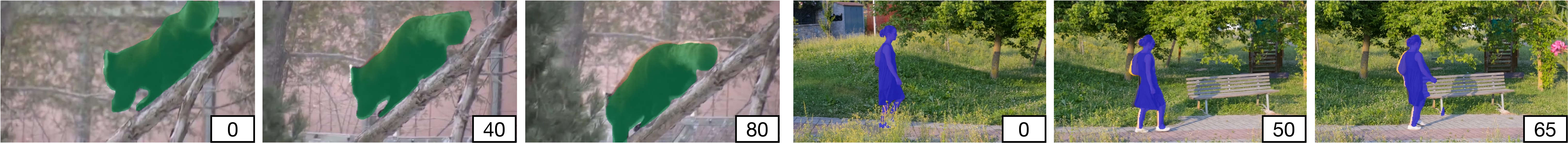} \\
\multicolumn{1}{c}{\footnotesize{(c) Qualitative results by introducing online adaptation}} \\
\end{tabular}
\caption[]{Qualitative comparisons on YouTube-VOS dataset~(left) and DAVIS-2017 train set~(right) for the effectiveness of the proposed modules.}
\label{fig:com_vis_2}
\end{center}
\end{figure*}

We show predictions from each separate proposed module in Figure~\ref{fig:com_vis_1} and \ref{fig:com_vis_2}.  
In Figure~\ref{fig:com_vis_1}, both the pose of the bird and the piglet vary a lot in the video. 
In addition, there is great appearance change in the bird object, 
yet our model is still able to produce good segmentation masks.
As shown in the visualisation results, 
the pairwise model tends to drift when the object has sharp pose variation or abrupt camera shaking between the paired frames. 
By adding the momentum memory, 
the model will compensate for the occlusion / dis-occlusion area in the previous frame by leveraging information from other previous frames, 
but the model is still prone to drift due to the fact that the intensities or colors of the foreground and background are highly similar. 
As shown in Figure~\ref{fig:com_vis_2}, online adaptation helps tackling this problem and clearing up some background drifts, 
as well as making segmentation masks smoother. 

\vspace{5pt}

\vspace{5pt}

\clearpage
\subsection{Visualisation of the online adaptation process.}
\begin{figure*}[!htb]
\footnotesize
\begin{center}
\begin{tabular}{c}
\includegraphics[width=1\textwidth]{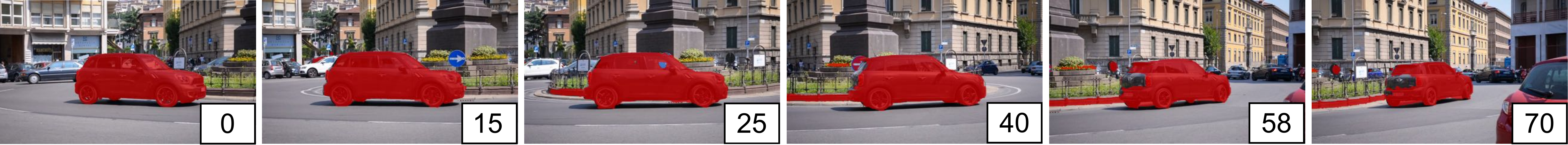} \\
\multicolumn{1}{c}{\footnotesize{(a) Propagated masks}} \\
\\ [-5pt]
\includegraphics[width=1\textwidth]{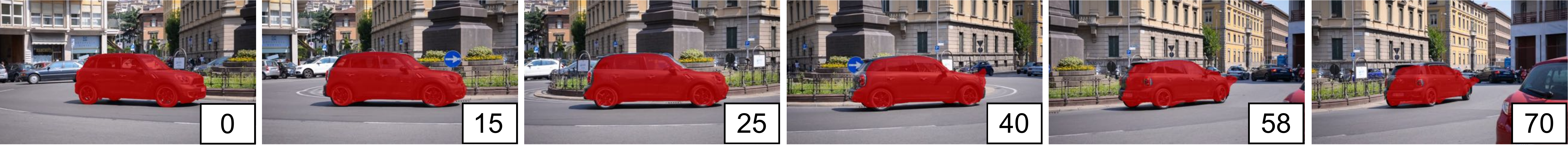} \\
\multicolumn{1}{c}{\footnotesize{(b) Segmentation results from appearance model of 100 training iterations}} \\
\\ [-5pt]
\includegraphics[width=1\textwidth]{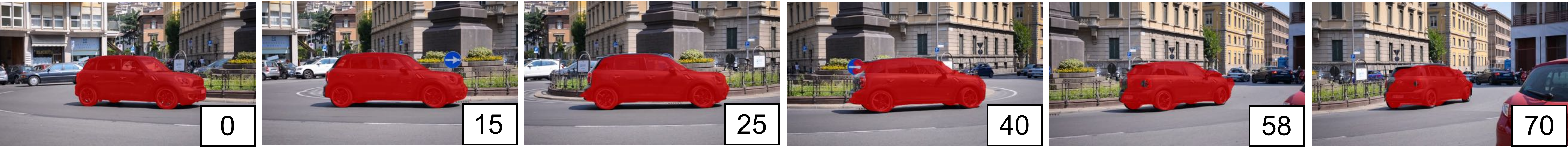} \\
\multicolumn{1}{c}{\footnotesize{(c) Segmentation results from appearance model of 200 training iterations}} \\
\\ [-5pt]
\includegraphics[width=1\textwidth]{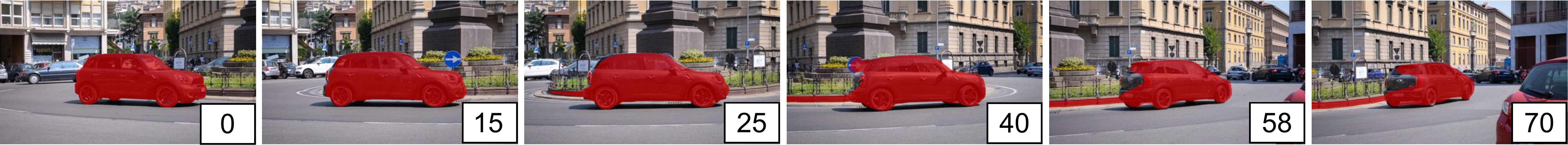} \\
\multicolumn{1}{c}{\footnotesize{(d) Segmentation results from appearance model of 300 training iterations}} \\
\end{tabular}
\caption[]{The visualisation results (DAVIS-2017) from the appearance model of different training iterations.}
\label{fig:ft_process_1}
\end{center}
\end{figure*}

\begin{figure*}[!htb]
\footnotesize
\begin{center}
\begin{tabular}{c}
\includegraphics[width=1\textwidth]{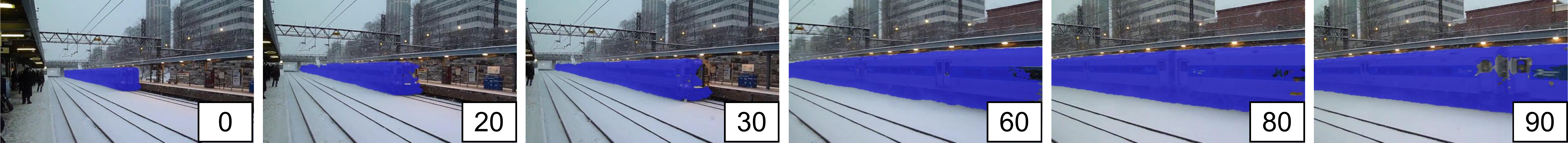} \\
\multicolumn{1}{c}{\footnotesize{(a) Propagated masks}} \\
\\ [-5pt]
\includegraphics[width=1\textwidth]{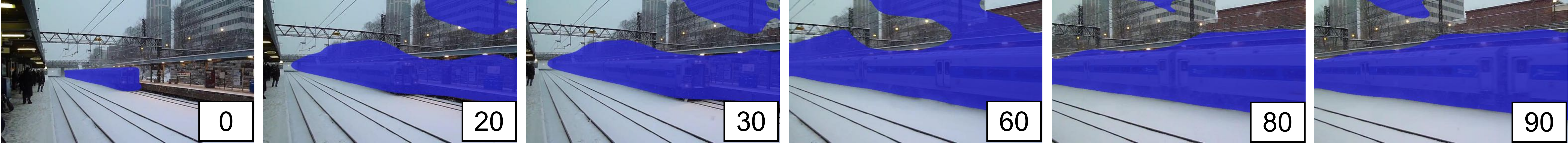} \\
\multicolumn{1}{c}{\footnotesize{(b) Segmentation results from appearance model of 50 training iterations}} \\
\\ [-5pt]
\includegraphics[width=1\textwidth]{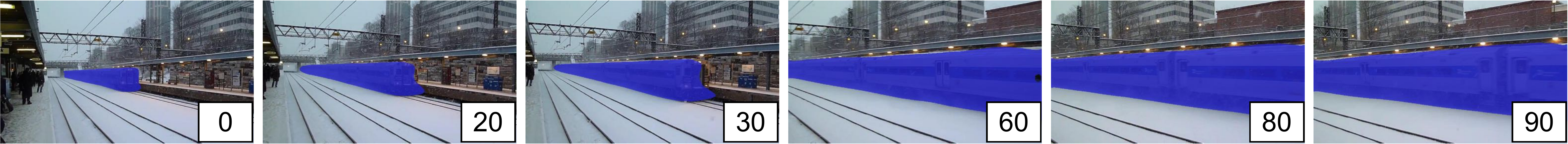} \\
\multicolumn{1}{c}{\footnotesize{(c) Segmentation results from appearance model of 100 training iterations}} \\
\\ [-5pt]
\includegraphics[width=1\textwidth]{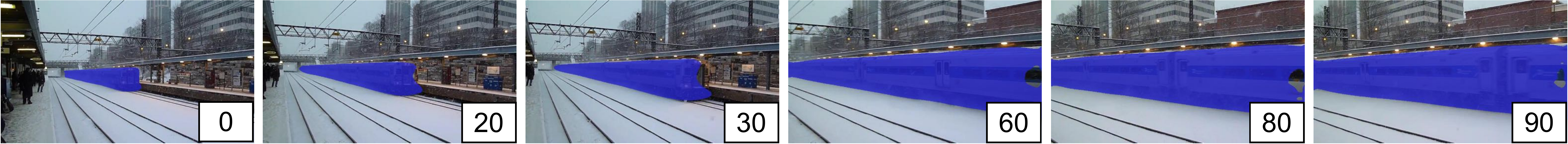} \\
\multicolumn{1}{c}{\footnotesize{(d) Segmentation results from appearance model of 200 training iterations}} \\
\end{tabular}
\caption[]{The visualisation results (YouTube-VOS) from the appearance model of different training iterations.}
\label{fig:ft_process_2}
\end{center}
\end{figure*}

We show the adapted results from the appearance model of different training iterations in Figure~\ref{fig:ft_process_1} and \ref{fig:ft_process_2}. The propagated masks used for training the appearance model contain drifts in the background. The figure indicates that during training, the appearance model first learns the self-similarity among object masks before fitting to the error drifts.

\end{document}